\documentclass[lettersize,journal]{IEEEtran}
\usepackage{amsmath,amsfonts}
\usepackage{array}
\usepackage[caption=false,font=normalsize,labelfont=sf,textfont=sf]{subfig}
\usepackage{CJKutf8}      
\usepackage{textcomp}
\usepackage{stfloats}
\usepackage{url}
\usepackage{verbatim}
\usepackage{graphicx}
\def\BibTeX{{\rm B\kern-.05em{\sc i\kern-.025em b}\kern-.08em
    T\kern-.1667em\lower.7ex\hbox{E}\kern-.125emX}}
\usepackage{balance}
\usepackage{booktabs} 
\usepackage{xcolor}         
\usepackage{enumitem} 
\definecolor{cvprblue}{rgb}{0.21,0.49,0.74}
\usepackage{multirow}
\usepackage[pagebackref,breaklinks,colorlinks,citecolor=cvprblue,urlcolor=cvprblue,linkcolor=cvprblue]{hyperref}
\usepackage{colortbl}
\usepackage{longtable}
\definecolor{Gray}{gray}{0.93}
\definecolor{Red}{RGB}{255, 46, 23}
\usepackage[ruled,vlined]{algorithm2e} 
\SetKwComment{Comment}{\tiny// }{}
\usepackage{amsmath} 
\usepackage{amssymb} 

\usepackage{booktabs} 
\usepackage{xcolor}         
\usepackage{pifont}
\definecolor{my_blue}{cmyk}{0.04, 0.02, 0, 0}
\definecolor{cvpr_blue}{cmyk}{0.72,0.34,0,0.26}
\definecolor{my_green}{cmyk}{0.04, 0, 0.06, 0.02}
\definecolor{acl_green}{cmyk}{0.74,0,0.21,0.40}
\definecolor{mycyan}{cmyk}{0.065, 0, 0, 0}
\usepackage{orcidlink} 

\begin{document}
\title{MCAT: Scaling Many-to-Many Speech-to-Text Translation \\ with MLLMs to 70 Languages}
\author{Yexing Du\orcidlink{0009-0003-0513-2635}, Kaiyuan Liu\orcidlink{0000-0001-7359-4450}, Youcheng Pan\orcidlink{0000-0002-8270-5455}, Bo Yang\orcidlink{0000-0002-4288-8349}, Keqi Deng\orcidlink{0000-0003-1490-963X}, \\Xie Chen\orcidlink{0000-0001-7423-617X}, Yang Xiang\orcidlink{0000-0003-1395-6805}, Ming Liu\orcidlink{0000-0001-7915-1001}, Bing Qin\orcidlink{0000-0002-2543-5604}, YaoWei Wang\orcidlink{0000-0002-6110-4036}

\thanks{
Manuscript received 2 December 2025; revised 24 February 2026; accepted 9 April 2026. Date of publication xx xxxx 2026; date of current version xx xxxx 2026. The research is supported by the National Key Research and Development Program of China (2025YFE0200500), the National Science Foundation of China (U22B2059, 62276083, 62506182), National Science and Technology Major Program (Grant No. 2024ZD01NL00101), the Key Research and Development Program of Heilongjiang Province (2022ZX01A28) and the 5G Application Innovation Joint Research Institute’s Project (A003), and the Major Key Project of PCL (Grant No. PCL2025A12, PCL2025A03). (\textit{Corresponding authors: Ming Liu; Yang Xiang.})

Yexing Du, Kaiyuan Liu and Yaowei Wang are with Harbin Institute of Technology, Shenzhen, China,
and also with Pengcheng Laboratory, Shenzhen,
China (e-mail: yxdu@ir.hit.edu.cn; 1171000408@stu.hit.edu.cn; 
wangyaowei@hit.edu.cn).

Ming Liu and Bing Qin are with
Harbin Institute of Technology, Harbin, China and also with Pengcheng Laboratory, Shenzhen,
China (e-mail: mliu@ir.hit.edu.cn; qinb@ir.hit.edu.cn). 

Youcheng Pan, Bo Yang and Yang Xiang are with Pengcheng Laboratory, Shenzhen, China (e-mail: panych@pcl.ac.cn; yangb05@pcl.ac.cn; xiangy@pcl.ac.cn).

Keqi Deng is with University of Cambridge, CB2 1TN Cambridge, U.K (e-mail: kd502@cam.ac.uk).

Xie Chen is with Shanghai Jiao Tong University, Shanghai, China (e-mail: chenxie95@sjtu.edu.cn).

}}

\markboth{Journal of \LaTeX\ Class Files,~Vol.~18, No.~9, September~2020}%
{How to Use the IEEEtran \LaTeX \ Templates}

\maketitle

\begin{abstract}
Multimodal Large Language Models (MLLMs) have achieved great success in Speech-to-Text Translation (S2TT) tasks. 
However, current research is constrained by two key challenges: language coverage and efficiency. 
Most of the popular S2TT datasets are substantially English-centric, which restricts the scaling-up of MLLMs' many-to-many translation capabilities. 
Moreover, the inference speed of MLLMs degrades dramatically when the speech is converted into long sequences (e.g., 750 tokens). 
To address these limitations, we propose a \underline{M}ultilingual \underline{C}ost-effective \underline{A}ccelerated Speech-to-Text \underline{T}ranslator (MCAT) framework, which includes two innovations. 
First, a language scaling method that leverages curriculum learning and a data balancing strategy are introduced to extend the language coverage supported by MLLMs to 70 languages and achieve mutual translation among these languages. 
Second, an optimized speech adapter module is designed to reduce the length of the speech sequence to only 30 tokens. 
Extensive experiments were conducted on MLLMs of different scales (9B and 27B). 
The experimental results demonstrate that MCAT not only surpasses state-of-the-art end-to-end models on the FLEURS dataset across $\mathbf{70 \times 69}$ directions but also enhances inference efficiency. 
\footnote{The code and models are released at \url{https://github.com/yxduir/m2m-70}.}
\end{abstract}

 
\section{Introduction}
Speech-to-Text Translation (S2TT) involves converting speech from a source language into text in a target language. Traditionally, S2TT tasks have relied on a cascaded system, where an Automatic Speech Recognition (ASR) model first transcribes the speech into text~\cite{baevski2020wav2vec}, followed by a Machine Translation (MT) model that translates the text into the target language~\cite{cheng2019breaking}. 
Recently, MLLMs~\cite{Qwen-Audio} have shown promise in simplifying model architectures and mitigating cascaded error propagation~\cite{sperber2020speech}.

\begin{figure}[t]
  \centering
  \includegraphics[width=\linewidth]{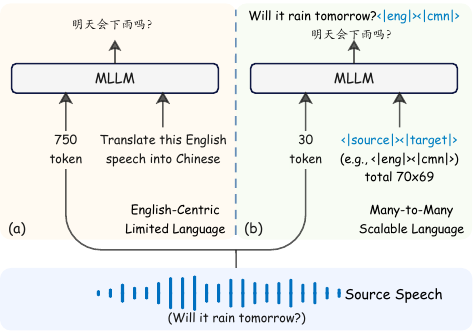} 
\caption{\textbf{Comparison of S2TT MLLMs.} (a) compresses speech to 750 tokens, has limited language support, and directly generates translated text; (b) generates transcriptions and translations in a single end-to-end pass, compressing speech to \textbf{30} tokens, supporting \textbf{70} languages. \texttt{<|eng|><|cmn|>} indicates transcribing English and translating it into Chinese.}
  \label{motivation}
\end{figure}

However, existing MLLMs for S2TT are constrained by two challenges: \textbf{language coverage} and \textbf{efficiency}. First, MLLM training is usually data-driven, but the existing S2TT datasets~\cite{wang2020covost} are predominantly English-centric. This leads to limited language coverage and weak many-to-many translation capabilities, as shown in Figure \ref{motivation}(a). Second, current MLLMs often employ an adapter structure similar to LLaVA~\cite{liu2023visual}, which uses an MLP directly to project features into the LLM, resulting in a very long input sequence (e.g., $\mathbf{750}$ tokens~\cite{Qwen-Audio}), even for extremely short samples such as ``Will it rain tomorrow?'', leading to limited inference efficiency.

To address these limitations, this research presents two key innovations. First, we introduce a \textbf{language scaling} strategy that includes a curriculum learning strategy (utilizing ASR data for pre-training, and minimal S2TT data to establish the connection between MT and S2TT), and a data balancing strategy to handle multilingual data imbalance. Finally, we extend the MLLM's S2TT task support to mutual translation among $\mathbf{70}$ languages, as shown in Figure~\ref{motivation}(b). Second, we design an efficient \textbf{speech adapter} structure, which utilizes a Q-Former~\cite{li2023blip} for feature extraction, pooling for compression, and an MLP for aligning the features to the LLM's dimension. This design reduces the speech token input to just \textbf{30 tokens}.

\begin{figure}[t]
  \centering
  \includegraphics[width=\linewidth]{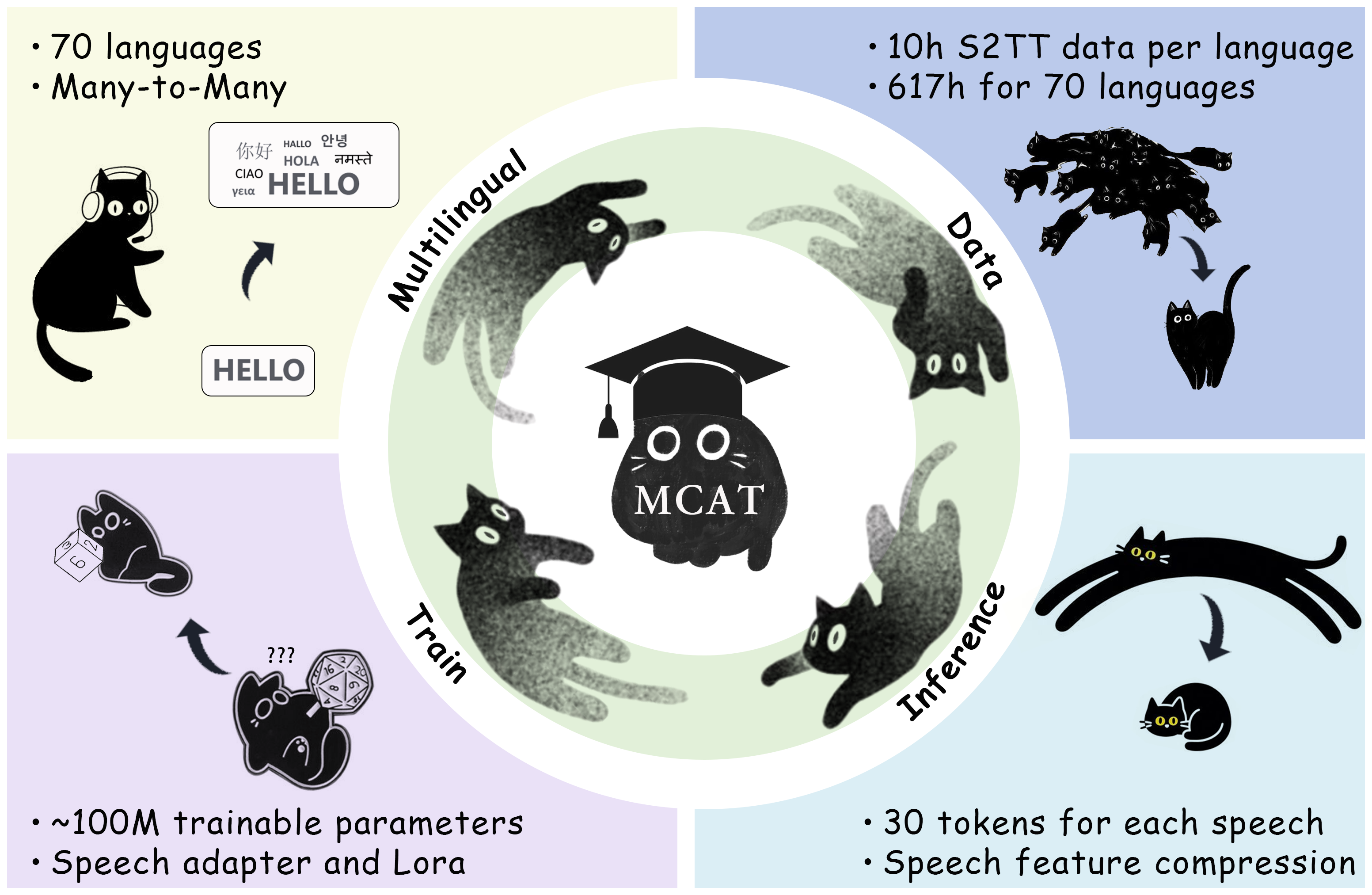} 
\caption{\textbf{Key Features:} (a) Multilingual Support; (b) Low-Resource Requirement; (c) Lightweight Training; (d) High-Efficiency Inference.}
  \label{MCAT}
\end{figure}

Based on the above design, \underline{M}ultilingual \underline{C}ost-effective \underline{A}ccelerated Speech-to-Text \underline{T}ranslator (MCAT) models exhibit four key features shown in Figure \ref{MCAT}: (1) Multilingual Support, offering many-to-many S2TT across $\mathbf{70}$ languages; (2) Low-Resource Requirement, needing only $\mathbf{10}$ hours of S2TT data per language; (3) Lightweight Training, achieved by utilizing the speech adapter and LoRA for efficient parameter training ($\sim\mathbf{100M}$ trainable parameters); and (4) High-Efficiency Inference, enabled by compressing the input speech sequences to just $\mathbf{30}$ tokens to accelerate batch inference.

To evaluate the impact of our proposed methods, we trained two MLLM variants of different scales ($\mathbf{9B}$ and $\mathbf{27B}$). Crucially, in low-resource settings, our models demonstrated powerful many-to-many S2TT capability on the FLEURS~\cite{fleurs2022arxiv} dataset across 70 languages, outperforming existing state-of-the-art end-to-end models.
Furthermore, we also validated the data scaling law on the CoVoST-2~\cite{wang2020covost} dataset. Finally, our strategies were validated by comprehensive ablation studies and comparisons of inference speed.

Our main contributions are summarized as follows:
\begin{itemize}
    \item We introduce a \textbf{Language Scaling Strategy} (including curriculum learning and data balancing) to enable many-to-many S2TT support across $\mathbf{70}$ languages, with comprehensive evaluation and analysis conducted across all $\mathbf{4,830}$ directions.
    \item We propose an optimized \textbf{Speech Adapter} that achieves an extreme $\mathbf{25\times}$ input compression, reducing the speech sequence length to $\mathbf{30}$ tokens. Despite such extreme compression, our model still achieved state-of-the-art end-to-end S2TT performance on FLEURS dataset.
    \item We validate that our MCAT framework is highly \textbf{Data- and Parameter-Efficient}. Our models ($\mathbf{9B}$ and $\mathbf{27B}$) achieve superior performance by fine-tuning only $\mathbf{\sim100M}$ parameters and utilizing minimal S2TT data ($\mathbf{<10}$ hours per language) for language extension.
\end{itemize}

In this paper, we extend our earlier work at ACL 2025~\cite{du2025making}. Specifically, we introduce a language scaling strategy to scale up the multilingual support from 15 to \textbf{70 languages}. Furthermore, we refine the speech adapter architecture to reduce the number of speech tokens to just \textbf{30}.

\newpage

\section{Related Work}

\subsection{Cascaded S2TT} The Cascaded S2TT approach typically employs a two-step pipeline: Automatic Speech Recognition (ASR) first transcribes the source spoken language into text, and subsequently Machine Translation (MT) translates the transcribed text into the target language. Specifically, established ASR models, such as Whisper~\cite{radford2023robust}, accurately convert speech into text. Similarly, MT models, for instance NLLB~\cite{nllb2024scaling}, achieve high translation accuracy and fluency by utilizing large multilingual datasets. However, a significant limitation of the cascaded approach is its susceptibility to error propagation.

\subsection{End-to-End S2TT} Distinct from the cascaded paradigm, End-to-End S2TT trains a unified model to directly map speech from the source language to text in the target language~\cite{sperber-etal-2019-attention,gaido-etal-2024-speech}, thereby eliminating the intermediate transcription step \cite{wang2020improving}. Certain models, like Whisper~\cite{radford2023robust}, also support multilingual-to-English translation capabilities. Furthermore, models such as SeamlessM4T-V2-Large~\cite{seamless2025joint} represent strong encoder-decoder architectures for diverse multilingual speech-to-text tasks. These pioneering efforts often prioritize reducing latency and enhancing efficiency over traditional offline speech translation systems.

\subsection{Audio MLLMs} Recently, the rapid advancements in MLLMs~\cite{li2025perception} have substantially improved performance in speech recognition and translation tasks. Approaches like SpeechGPT~\cite{zhang2023speechgpt} utilize prompting mechanisms to enhance speech recognition within large language models. SALMONN~\cite{tang2023salmonn} specifically focuses on improving the auditory comprehension of both language and music. Qwen-Audio~\cite{Qwen-Audio} advances audio recognition and translation by retraining speech encoders within a multi-task framework. More recently, Voxtral~\cite{liu2025voxtral} and Qwen3-Omni~\cite{xu2025qwen3} have further extended this progress by integrating enhanced multimodal understanding. In addition, LLM-SRT~\cite{du2025making} introduces a curriculum learning strategy designed to strengthen cross-modal alignment and translation quality.

\begin{table}[b]
    \centering
    \small
    \caption{S2TT Language Coverage.}
    \label{tab:summary}
      \setlength{\tabcolsep}{4pt} 
    \renewcommand{\arraystretch}{1.1} 
    \resizebox{1.0\linewidth}{!}{
    \begin{tabular}{lccc}
        \toprule
      \textbf{S2TT Models} &\textbf{Language}      &\textbf{S2TT Data (h)}             \\
         \midrule
        \multicolumn{3}{c}{\textbf{Encoder-Decoder Models}} \\ \midrule
             \multirow{1}{*}{Whisper-Large-V2~\cite{radford2023robust}}   &96 $\rightarrow$ eng                                   &   125,000 \\
             \multirow{1}{*}{SeamlessM4T-V2-Large~\cite{seamless2025joint}}   &101 $\leftrightarrow$ 96                                  &   351,000 \\

        \midrule
        \multicolumn{3}{c}{\textbf{MLLMs}} \\ \midrule
        \multirow{1}{*}{Qwen-Audio-7B~\cite{Qwen-Audio}}   &6 $\leftrightarrow$ 6                                   & 3,700   \\
             \multirow{1}{*}{Voxtral-Small-24B}~\cite{liu2025voxtral}   &8 $\leftrightarrow$ 8                                  &  in-house  \\
     \multirow{1}{*}{Qwen3-Omni-30B-A3B-Instruct~\cite{xu2025qwen3}}   &19 $\leftrightarrow$ 19                                   &   in-house \\
        \multirow{1}{*}{MCAT-Small-9B (ours)}   &28 $\leftrightarrow$ 28                                   &  243.9  \\
     \multirow{1}{*}{MCAT-Large-27B (ours)}   &70 $\leftrightarrow$ 70                                   &   617.7 \\

        \bottomrule
    \end{tabular}}
    \raggedright{\hspace*{0.5em} Language coverage follows reported S2TT scores in the paper.}
\end{table}

\newpage

\section{Methodology}

\subsection{Model Architecture}
\label{sec:3.1}
As detailed in Table \ref{tab:parameters}, the MCAT models are built upon an LLM. They adopt Whisper's encoder~\cite{radford2023robust} as the speech encoder, followed by a Q-Former~\cite{li2023blip}, Pooling, and MLP layer for the speech adapter. Notably, our design compresses \textbf{30 seconds of speech into 30 tokens} to boost MLLM inference efficiency. 

\subsubsection{\textbf{Speech Preprocessing}}
The raw waveform $x \in \mathbb{R}^{N \times T}$ ($N$ being the batch size and $T$ the temporal length) undergoes audio processing, including STFT and Mel Filterbanks, to convert the time-domain signal into a Mel-spectrogram $M$.
\begin{equation}
x \in \mathbb{R}^{N \times T}
\quad \xrightarrow[\text{STFT}]{\text{Mel Filterbanks}}
\quad M \in \mathbb{R}^{N \times C \times L},
\end{equation}
where the Whisper encoder requires the input to be a fixed length $L$, achieved by truncation or padding. The dimension $C$ represents the key feature size of the Mel-spectrogram.

\subsubsection{\textbf{Speech Encoder}}

We leverage the frozen Whisper's encoder, which maps the padded Mel-spectrogram input $\mathbf{X}$ to a sequence of hidden representations $\mathbf{H}$:
\begin{equation}
H = \text{Encoder}(M),\quad H \in \mathbb{R}^{N \times L' \times D_w}
\end{equation}
where $D_w$ is the encoder hidden dimension and $L'$ is the encoder sequence length.

\begin{table}[b]
    \centering
    \small
    \caption{Stages and Output Shapes.}
    \label{tab:stage}
    \setlength{\tabcolsep}{3pt} 
    \renewcommand{\arraystretch}{1.5} 
    \resizebox{1.0\linewidth}{!}{
    \begin{tabular}{lllc}
        \toprule
        \textbf{Input} & \textbf{Stage} & \textbf{Feature} & \textbf{Shape} \\
        \midrule
        \multirow{6}{*}{Speech} & Raw Speech & $x \in \mathbb{R}^{N \times T}$ & $N \times T$ \\
        & Mel-spectrogram & $M \in \mathbb{R}^{N \times C \times L}$ & $N \times 128 \times 3000$ \\
        & Speech Encoder & $H \in \mathbb{R}^{N \times L' \times D_w}$ & $N \times 1500 \times 1280$ \\
        & Q-Former & $Z \in \mathbb{R}^{N \times K \times D_q}$ & $N \times 150 \times 768$ \\
        & Pooling & $Z_p \in \mathbb{R}^{N \times K/S \times D_q}$ & $N \times 30 \times 768$ \\
        & MLP & $Z_{mlp} \in \mathbb{R}^{N \times K/S \times D_{llm}}$ & $N \times 30 \times D_{llm}$ \\ 
        \midrule
        Text & Text Embedding & $P \in \mathbb{R}^{N \times P_t \times D_{llm}}$ & $N \times P_t \times D_{llm}$ \\ 
        \midrule
        LLM Input & Multimodal Fusion & $X \in \mathbb{R}^{N \times (K/S + P_t) \times D_{llm}}$ & $N \times (30 + P_t) \times D_{llm}$ \\
        \bottomrule
    \end{tabular}}
\end{table}

\subsubsection{\textbf{Speech Adapter}}
The speech adapter layer comprises a Q-Former, a pooling layer, and an MLP. The Q-Former is responsible for feature extraction, the pooling layer handles feature compression, and the MLP layer performs dimension alignment with the LLM's embedding space.
\paragraph{\textbf{Q-Former for Feature Extraction}}
The Q-Former serves to extract a set of compact, fixed-length speech embeddings $Z$ from the longer sequence $H$:
\begin{equation}
Z = \text{Q-Former}(H), \quad Z \in \mathbb{R}^{N \times K \times D_q}
\end{equation}
Here, $K$ is the fixed number of learned query tokens, and $D_q$ is the Q-Former's hidden dimension.

\paragraph{\textbf{Pooling Layer for Feature Compression}}
Temporal pooling is applied to reduce the sequence length by $S\times$ downsampling (e.g., average pooling):
\begin{equation}
Z_p = \text{Pool}(Z), \quad Z_p \in \mathbb{R}^{N \times K/S \times D_q}
\end{equation}
This operation compresses the $K$ acoustic features to $K/S$.

\paragraph{\textbf{MLP for Dimension Alignment}}
The MLP  maps the compressed features into the LLM's dimension $D_{llm}$:
\begin{equation}
Z_{mlp} = \text{MLP}(Z_p), \quad Z_{mlp} \in \mathbb{R}^{N \times K/S \times D_{llm}}
\end{equation}
where $Z_{mlp}$ represents the aligned speech feature embeddings, ready for concatenation.

\subsubsection{\textbf{Text Embedding}}
Given the instruction text $t$, the corresponding prompt embeddings are obtained as:
\begin{equation}
P = \text{Embedding}(t) \in \mathbb{R}^{N \times P_t \times D_{llm}},
\end{equation}
where $P_t$ is the prompt token length.

\subsubsection{\textbf{Multimodal Fusion and LLM Output}}

To achieve multimodal integration, the modality-specific features $Z_{mlp}$ are fused with the text embeddings $P$ by concatenating them along the temporal dimension:

\begin{equation}
X = \text{Concat}(Z_{mlp}, P) \in \mathbb{R}^{N \times (K/S + P_t) \times D_{llm}}
\end{equation}

The fused representation $X$ is subsequently fed into the LLM, which autoregressively produces the text outputs $Y$.

\begin{figure}[t]
\centering
 \includegraphics[width=0.95\linewidth]{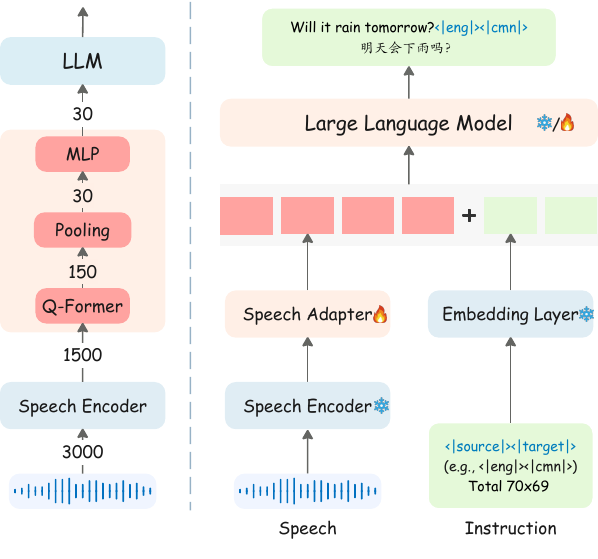}
\caption{\textbf{The Architecture of MCAT Model.} Our MLLM compresses the input audio into 30 tokens, supporting a total of 70 languages.}
\label{framework}
\end{figure}

\begingroup

\begin{table*}[t]
  \centering
  \small
  \renewcommand{\arraystretch}{1.3} 
  \caption{Instruction Design.}
  \setlength{\tabcolsep}{6pt} 
    \resizebox{1.0\linewidth}{!}{
    \begin{tabular}{c c c c} 
    \toprule 

\textbf{Task}&\textbf{Speech}&\textbf{Instruction Text }&\textbf{Prediction} \\ 
\hline

\multirow{2}{*}{ASR} &{\color{cvpr_blue}\ding{51}} &\texttt{<|eng|>}  & Will it rain tomorrow? \\  
 &{\color{cvpr_blue}\ding{51}} &\texttt{<|deu|>}  &  Regnet es morgen?	\\  \hline

\multirow{2}{*}{SMT} &{\color{cvpr_blue}\ding{51}} &Will it rain tomorrow?\texttt{<|eng|><|deu|>}   & Regnet es morgen?    \\ 
 &{\color{cvpr_blue}\ding{51}} &Regnet es morgen?\texttt{<|deu|><|fra|>}  & Il va pleuvoir demain?  \\ \hline

\multirow{2}{*}{SRT} &{\color{cvpr_blue}\ding{51}}&\texttt{<|eng|><|deu|>} & Will it rain tomorrow?\texttt{<|eng|><|deu|>}Regnet es morgen?   \\ 
 &{\color{cvpr_blue}\ding{51}} &\texttt{<|deu|><|fra|>}  & Regnet es morgen?\texttt{<|deu|><|fra|>}Il va pleuvoir demain ?  \\

\bottomrule
    \end{tabular}}
\label{pattern}
\end{table*}%
\endgroup

\subsection{Task Formulation}
\label{sec:3.0}
In this section, we define the following tasks: 
\begin{itemize}[itemsep=2ex] 
\item \textbf{Automatic Speech Recognition (ASR)}: Given the speech input \( x \) and the instruction text \( t \), the goal is to produce the transcribed text \( Y_1 \).

\item \textbf{Speech-guided Machine Translation (SMT)}: Given the speech input \( x \), its transcription \( Y_1 \), and the instruction text \( t \), the goal is to produce the translated text \( Y_2 \).

\item \textbf{Speech Recognition and Translation (SRT)}: Given the speech input \( x \) and the instruction text \( t \), the goal is to produce the transcription \( Y_1 \) and the translation \( Y_2 \).
\end{itemize}

\newpage

\subsection{Language Scaling Strategy}

\label{sec:3}

MCAT employs a comprehensive language scaling strategy to effectively train an MLLM for multilingual S2TT across $70$ languages. This strategy involves a three-stage curriculum learning strategy to bridge the connection between MT and S2TT tasks and a data balancing strategy focusing on balanced ASR and S2TT data usage.

\subsubsection{\textbf{Language Tags}} 
Minimalist instructions are designed to help the model distinguish between tasks while minimizing instruction token length, as shown in Table~\ref{pattern}. This design ensures that task-specific markers, such as \texttt{<|eng|><|deu|>}, appear in the generated answers, effectively segmenting transcription and translation content in the SRT task.

\subsubsection{\textbf{Curriculum Learning Strategy}}
\label{sec:3.3}

We adopt a curriculum learning approach that incorporates three sequential training tasks: ASR, SMT, and SRT. This sequence is designed to utilize data-rich ASR as a bridge to develop fundamental capabilities before scaling to the more complex SMT and SRT tasks.

\paragraph{\textbf{ASR Pre-training}} 
In this initial stage, the model is pre-trained to develop ASR capabilities with a focus on multimodal alignment. This step also involves expanding language support by training on all intended languages. The speech adapter is trained with as much data as possible to ensure efficient fine-tuning and establish a strong foundation.

\paragraph{\textbf{SMT Enhancement}} 
This stage enhances the model's cross-lingual abilities. Starting from the ASR checkpoint, the model takes both transcribed text and audio as input to generate translations based on the instruction. The purpose is to activate the LLM's inherent MT capabilities and establish the necessary connection between the MT and the S2TT tasks.

\paragraph{\textbf{SRT Activation}} 
The final stage activates the model's full SRT capabilities. Training continues from the SMT checkpoint, with the model receiving only audio input and a task-specific instruction, outputting both the transcription and translation of the speech. This step extends the MT capabilities of LLMs to the S2TT task and finalizes the model.

\begin{algorithm}[t]
\small
\caption{Language Scaling Strategy}
\label{alg:language_scaling_strategy}
\KwIn{Initial Model $\mathcal{M}_0$}
\KwOut{Final Model $\mathcal{M}_{\text{final}}$}
Let $\mathcal{D}$ be the training dataset
\BlankLine
\textbf{Phase 1: ASR Pre-training} 

\Comment{Activate ASR Capabilities}
\label{alg:phase1}
\Begin{
    
$\mathcal{L}_{\text{set}} \leftarrow \langle \mathcal{L}_2, \mathcal{L}_{28}, \mathcal{L}_{44}, \mathcal{L}_{\text{full}} \rangle$\

\Comment{Language Sets}
    \For{$\mathcal{L}_{\text{subset}}$ in $\mathcal{L}_{\text{set}}$}
    {
        $\mathcal{D}_{\text{ASRsubset}} \leftarrow \text{ASRdata}(\mathcal{L}_{\text{subset}})$ 
        
        $\mathcal{M}_{\text{ASR}} \leftarrow \text{SFT}
        (\mathcal{M}_{\text{0}}, \mathcal{D}_{\text{ASRsubset}})$ 

        \Comment{Audio $\to$ Transcription}
    }
}
\textbf{Phase 2: Balanced ASR Fine-Tuning} \label{alg:phase2}
  
    \Comment{Balanced samples per language}

\Begin{
    $\mathcal{D}_{\text{BalancedASR}} \leftarrow \text{ASRdata}(\mathcal{L}_{\text{full}}, \text{Max}=10000)$\
    
    $\mathcal{M}_{\text{BalancedASR}} \leftarrow \text{SFT}(\mathcal{M}_{\text{ASR}}, \mathcal{D}_{\text{BalancedASR}})$\
}
\textbf{Phase 3: SMT and SRT Fine-Tuning}

\Comment{Activate S2TT Capabilities}
\label{alg:phase3}
\Begin{
    $\mathcal{D}_{\text{SMT}} \leftarrow \text{SMTdata}(\mathcal{L}_{\text{full}})$\
    
    $\mathcal{M}_{\text{SMT}} \leftarrow \text{SFT}(\mathcal{M}_{\text{BalancedASR}}, \mathcal{D}_{\text{SMT}})$ 
    
    \Comment{Audio + Transcription $\to$ Translation}

    $\mathcal{D}_{\text{SRT}} \leftarrow \text{SRTdata}(\mathcal{L}_{\text{full}})$\
    
    $\mathcal{M}_{\text{SRT}} \leftarrow \text{SFT}(\mathcal{M}_{\text{SMT}}, \mathcal{D}_{\text{SRT}})$ 
    
    \Comment{Audio $\to$ Transcription + Translation}
}
\textbf{Phase 4: Balanced SRT Fine-Tuning} 

        \Comment{Balanced samples per direction}

\label{alg:phase4}
\Begin{
    $\mathcal{D}_{\text{BalancedSRT}} \leftarrow \text{SRTdata}(\mathcal{L}_{\text{full}}, \text{Max}=100)$ 
    
    $\mathcal{M}_{\text{final}} \leftarrow \text{SFT}(\mathcal{M}_{\text{SRT}}, \mathcal{D}_{\text{BalancedSRT}})$ 
}
\end{algorithm}

\subsubsection{\textbf{Data Balancing Strategy}}
\label{sec:3.2}
A core challenge when training our $70$-language model is mitigating disparity in performance caused by inherent data imbalance. We employ a strategy that scales the language set from high-resource to low-resource languages, followed by a final balancing step.

\paragraph{\textbf{ASR Language Expansion}}
Training begins with English and Chinese $\text{ASR}$ data for foundational capability. The language set is then progressively expanded in stages: first to $28$ languages, then to $44$, and finally to the full $70$ languages.

\paragraph{\textbf{Balanced ASR Fine-Tuning}}
In this stage, we reduced the ASR data for all languages to a maximum of 10,000 samples per language, and then continued ASR training based on the previous checkpoint.

\paragraph{\textbf{SMT and SRT Full-Scale Training}}
We continue training from the ASR checkpoint using data from all $\mathbf{70}$ languages to enhance S2TT capabilities. The model is first fine-tuned on the SMT task, then on the SRT task.

\paragraph{\textbf{Balanced SRT Fine-Tuning}}
In this stage, we reduced the SRT data for all language directions to a maximum of 100 samples per direction, and then continued SRT training based on the previous checkpoint.

\newpage

\section{Experiments Setting}

\subsection{Datasets}
In our experiments, we use the \textit{CommonVoice}\footnote{\url{https://datacollective.mozillafoundation.org/datasets?q=common+voice}}~\cite{commonvoice:2020} and \textit{FLEURS}\footnote{\url{https://huggingface.co/datasets/google/fleurs}}~\cite{fleurs2022arxiv} datasets for the \textbf{ASR} task training, and the \textit{FLEURS} dataset for the \textbf{SMT} and \textbf{SRT} task training. We perform comparative and ablation studies on the \textit{FLEURS} and \textit{CoVoST-2}\footnote{\url{https://github.com/facebookresearch/covost}}~\cite{wang2020covost} datasets. Detailed information for datasets is provided in Table \ref{tab:dataset}.

\subsection{Model Architecture} As shown in Table \ref{tab:parameters}, the MLLM consists of an LLM (GemmaX2-9B~\cite{cui2025multilingual} or Gemma3-27b-it~\cite{team2025gemma}), a frozen speech encoder (Whisper-large-v3), and a trainable adapter layer comprising a Q-Former, Pooling layer and MLP layer. For Q-Former, we use 150 queries, each with a dimension of 768. Training can be minimized by freezing the LLM, or LoRA \cite{hu2021lora} can be applied for training.

\subsection{Training Details}
We used BF16 precision with Distributed Data Parallel (DDP), a learning rate of \(5 \times 10^{-5}\), 1000 warmup steps, and the AdamW optimizer. The models were trained on 8 A100 GPUs. 
The 9B model can be trained in 3 days, while the 27B model can be trained in 7 days.

\subsection{Compared Methods}
We compare both cascade systems and end-to-end S2TT models, such as  SeamlessM4T~\cite{barrault2023seamlessm4t} which supports S2TT for nearly 100 languages, and Qwen-Omni series~\cite{xu2025qwen3}, the open-source MLLM that centers on English and Chinese, extending its capabilities to diverse audio modalities.

\subsection{Language Support}
As shown in Table \ref{language}, our MCAT-Small model supports 28 languages across 9 language families, while the MCAT-Large model supports 70 languages across 12 language families. Instructions for language support can be found in the appendix.

\begin{table}[h] 
\centering 
\footnotesize 
\caption{MLLM Training Settings. }
\renewcommand{\arraystretch}{1.2} 
\setlength{\tabcolsep}{1mm} 
\resizebox{1.0\linewidth}{!}{
\begin{tabular}{lcccl} 
\toprule
\multirow{2}{*}{\textbf{Modules}} & \multirow{1}{*}{\textbf{MCAT}}& \multirow{1}{*}{\textbf{MCAT}} & \textbf{Train} & \multirow{2}{*}{\textbf{Details}} \\
& \textbf{-Small}&\textbf{-Large} &\textbf{Stage} & \\
\midrule
Speech Encoder & $\sim$635M & $\sim$635M & - & Whisper's encoder \\
\midrule
Speech Adapter & {\color{cvpr_blue}$\sim$80.6M}& {\color{cvpr_blue}$\sim$85.2M} & All & Q-Former / Pooling / MLP \\
\midrule
LLM & $\sim$9.2B & $\sim$27.4B & -& GemmaX2-9B or Gemma3-27B\\
\midrule
LLM Lora & {\color{cvpr_blue}$\sim$8.9M} & {\color{cvpr_blue}$\sim$18.7M}& SRT & LoRA (r=16, alpha=32) \\ \midrule
Total Trainable& {\color{cvpr_blue}$\sim$89.5M} & {\color{cvpr_blue}$\sim$103.9M} & & \\ \midrule
Total & $\sim$10B& $\sim$28B& & \\ 
\bottomrule
\end{tabular}}
\raggedright{\hspace*{0.5em}The {\color{cvpr_blue}blue} color indicates the trainable parameters.}
\label{tab:parameters} 
\end{table}

\newpage

\begin{table}[t] 
\centering
\tiny
\caption{Language Support.}
\label{tab:taslp71_full_languages}
\setlength{\tabcolsep}{6pt} 
\renewcommand{\arraystretch}{0.85} 
\resizebox{1.0\linewidth}{!}{
\begin{tabular}{c c c c c c} 
\toprule
\textbf{ISO } & \multirow{2}{*}{\textbf{Language}} & \multirow{2}{*}{\textbf{Family}}& {\textbf{MCAT}}& {\textbf{MCAT}}& \textbf{S2TT } \\
\textbf{Code} &  & &\textbf{-Small} & \textbf{-Large}& \textbf{Data (h)} \\
\midrule
afr & Afrikaans & Indo-European&&\ding{51} 					&	3.6 	\\
amh & Amharic & Afro-Asiatic&&\ding{51} 					&	11.1 	\\
ara & Arabic & Afro-Asiatic &\ding{51}&\ding{51}					&	6.0 	\\
asm & Assamese & Indo-European &&\ding{51}					&	10.7 	\\
azj & Azerbaijani & Turkic&&\ding{51} 					&	9.3 	\\
bel & Belarusian & Indo-European &&\ding{51}					&	9.5 	\\
ben & Bengali & Indo-European&\ding{51} &\ding{51}					&	10.7 	\\
bos & Bosnian & Indo-European &&\ding{51}					&	10.0 	\\
bul & Bulgarian & Indo-European&&\ding{51} 					&	9.5 	\\
cat & Catalan & Indo-European&&\ding{51} 					&	7.4 	\\
ces & Czech & Indo-European&\ding{51} &\ding{51}					&	8.4 	\\
cmn & Chinese & Sino-Tibetan &\ding{51}&\ding{51}					&	9.7 	\\
cym & Welsh & Indo-European &&\ding{51}					&	12.2 	\\
dan & Danish & Indo-European&&\ding{51}					&	7.5 	\\
deu & German & Indo-European&\ding{51} &\ding{51} 					&	9.0 	\\
ell & Greek & Indo-European&&\ding{51} 					&	10.0 	\\
eng & English & Indo-European&\ding{51}&\ding{51} 					&	7.5 	\\
est & Estonian & Uralic&&\ding{51} 					&	7.3 	\\
fas & Persian & Indo-European &\ding{51}&\ding{51}					&	12.1 	\\
fin & Finnish & Uralic &&\ding{51}					&	8.8 	\\
fra & French & Indo-European &\ding{51}&\ding{51}					&	10.3 	\\
glg & Galician & Indo-European&&\ding{51} 					&	6.7 	\\
guj & Gujarati & Indo-European&&\ding{51} 					&	9.0 	\\
heb & Hebrew & Afro-Asiatic &\ding{51}&\ding{51}					&	9.5 	\\
hin & Hindi & Indo-European &\ding{51}&\ding{51}					&	6.7 	\\
hrv & Croatian & Indo-European &&\ding{51}					&	11.8 	\\
hun & Hungarian & Uralic&&\ding{51} 					&	9.3 	\\
hye & Armenian & Indo-European &&\ding{51}					&	10.4 	\\
ind & Indonesian & Austronesian &\ding{51}&\ding{51}					&	9.1 	\\
isl & Icelandic & Indo-European&&\ding{51} 					&	2.8 	\\
ita & Italian & Indo-European &\ding{51}&\ding{51}					&	9.0 	\\
jav & Javanese & Austronesian&&\ding{51} 					&	11.2 	\\
jpn & Japanese & Japonic &\ding{51}&\ding{51}					&	7.4 	\\
kan & Kannada & Dravidian &&\ding{51}					&	8.3 	\\
kat & Georgian & Kartvelian &&\ding{51}					&	5.1 	\\
kaz & Kazakh & Turkic &&\ding{51}					&	11.8 	\\
khm & Khmer & Austroasiatic &\ding{51}&\ding{51}					&	7.1 	\\
kir & Kyrgyz & Turkic &&\ding{51}					&	9.3 	\\
kor & Korean & Koreanic &\ding{51}&\ding{51}					&	7.9 	\\
lao & Lao & Kra–Dai &\ding{51}&\ding{51}					&	7.3 	\\
lav & Latvian & Indo-European &&\ding{51}					&	6.5 	\\
lit & Lithuanian & Indo-European &&\ding{51}					&	9.8 	\\
mal & Malayalam & Dravidian &&\ding{51}					&	10.1 	\\
mkd & Macedonian & Indo-European &&\ding{51}					&	6.8 	\\
msa & Malay & Austronesian &\ding{51}&\ding{51}					&	9.5 	\\
mya & Burmese & Sino-Tibetan &\ding{51}&\ding{51}					&	12.1 	\\
nld & Dutch & Indo-European &\ding{51}&\ding{51}					&	7.7 	\\
nob & Norwegian & Indo-European &&\ding{51}					&	10.9 	\\
npi & Nepali & Indo-European &&\ding{51}					&	11.3 	\\
pan & Punjabi & Indo-European &&\ding{51}					&	6.4 	\\
pol & Polish & Indo-European &\ding{51}&\ding{51}					&	9.2 	\\
por & Portuguese & Indo-European &\ding{51}&\ding{51}					&	10.2 	\\
ron & Romanian & Indo-European &&\ding{51}					&	10.1 	\\
rus & Russian & Indo-European &\ding{51}&\ding{51}					&	8.1 	\\
slk & Slovak & Indo-European &&\ding{51}					&	5.9 	\\
slv & Slovenian & Indo-European &&\ding{51}					&	7.8 	\\
spa & Spanish & Indo-European &\ding{51}&\ding{51}					&	8.8 	\\
srp & Serbian & Indo-European &&\ding{51}					&	10.7 	\\
swe & Swedish & Indo-European &&\ding{51}					&	8.4 	\\
swh & Swahili & Niger–Congo &&\ding{51}					&	13.5 	\\
tam & Tamil & Dravidian &&\ding{51}					&	8.7 	\\
tel & Telugu & Dravidian &&\ding{51}					&	7.9 	\\
tgl & Tagalog & Austronesian &\ding{51}&\ding{51}					&	7.7 	\\
tha & Thai & Kra–Dai &\ding{51}&\ding{51}					&	8.5 	\\
tur & Turkish & Turkic &\ding{51}&\ding{51}					&	8.3 	\\
ukr & Ukrainian & Indo-European &&\ding{51}					&	9.0 	\\
urd & Urdu & Indo-European &\ding{51}&\ding{51} 					&	7.0 	\\
uzb & Uzbek & Turkic &&\ding{51}					&	10.1 	\\
vie & Vietnamese & Austroasiatic &\ding{51}&\ding{51}					&	9.1 	\\
yue & Cantonese & Sino-Tibetan &&\ding{51}					&	7.3 	\\ \midrule
Total& &&28&70& 617.7\\
\bottomrule
\end{tabular}}
\label{language}
\end{table}

\subsection{Evaluation Metrics.} We use COMET\footnote{\url{https://unbabel.github.io/COMET/}}~\cite{rei2022comet} and spBLEU\footnote{\url{https://github.com/mjpost/sacrebleu}}~\cite{post-2018-call} as evaluation metrics. The spBLEU utilizes the FLORES-200 tokenizer. Instructions for evaluation metrics can be found in the appendix.

\newpage

\newpage

\begingroup
\renewcommand{\arraystretch}{1.1} 
        \setlength{\tabcolsep}{6pt} 
\begin{table*}[!ht]
\centering
\small
\caption{COMET Results on $9\times27$ and $9\times69$ Directions on the FLEURS Dataset. spBLEU Results are shown in Table \ref{0970_spbleu}.
}
\resizebox{1.0\textwidth}{!}{
\begin{tabular}{l|ccccccccc|c}
\toprule
\rowcolor{gray!12}
  &                     &                    &                    &                   &                    &      &                   &                     &                    &                                        \\
\rowcolor{gray!12}
\multirow{-2}{*}{\textbf{Systems} (X $\rightarrow$ 27)}                                                      &   \multirow{-2}{*}{ara}              &    \multirow{-2}{*}{cmn}              &    \multirow{-2}{*}{eng}                       &  \multirow{-2}{*}{ind}               &       \multirow{-2}{*}{jpn}           &   \multirow{-2}{*}{kor}                        &         \multirow{-2}{*}{tha}         &         \multirow{-2}{*}{tur}        &       \multirow{-2}{*}{vie}    &          \multirow{-2}{*}{\textbf{Avg.}}                    \\
\midrule
\multicolumn{11}{c}{\textbf{Cascaded ASR+MT Models}} \\
\midrule
Whisper-Large-V3 + NLLB-200-3.3B~\cite{nllb2024scaling}&\underline{78.1}&\underline{79.8}&83.4&\underline{81.6}&\underline{79.9}&\underline{81.2}&\underline{78.2}&\underline{82.3}&\underline{78.3}&\underline{80.3}               \\   
Whisper-Large-V3 + LLaMAX3-8B-Alpaca~\cite{lu2024llamax}  &   76.0&78.6&81.3&79.4&78.4&79.3&77.0&79.2&77.1&78.5          \\  
\midrule
\multicolumn{11}{c}{\textbf{End-to-end S2TT Models}} \\
\midrule
SeamlessM4T-V2-Large~\cite{seamless2025joint} &     69.4&72.6&\underline{84.3}&71.2&69.1&73.5&68.6&71.2&71.9&72.4    \\
MCAT-Small-9B (ours)    &{77.8}&\colorbox{cvprblue!14}{79.8}&\colorbox{cvprblue!14}{85.9}&\colorbox{cvprblue!14}{82.3}&79.7&\colorbox{cvprblue!14}{81.9}&\colorbox{cvprblue!14}{\textbf{78.6}}&82.1&\colorbox{cvprblue!14}{79.3}&\colorbox{cvprblue!14}{80.8}        \\
MCAT-Large-27B (ours) &     \colorbox{cvprblue!14}{\textbf{78.7}}&\colorbox{cvprblue!14}{\textbf{80.3}}&\colorbox{cvprblue!14}{\textbf{86.3}}&\colorbox{cvprblue!14}{\textbf{83.2}}&79.8&\colorbox{cvprblue!14}{\textbf{82.4}}&78.0&\colorbox{cvprblue!14}{\textbf{83.2}}&\colorbox{cvprblue!14}{\textbf{79.5}}&\colorbox{cvprblue!14}{\textbf{81.3}}    \\
\toprule

\rowcolor{gray!12}
   &                     &                     &                     &                    &                   &   &                   &                    &                   &                           \\
\rowcolor{gray!12}
\multirow{-2}{*}{\textbf{Systems} (X $\rightarrow$ 69)}                        &   \multirow{-2}{*}{ara}              &    \multirow{-2}{*}{cmn}              &    \multirow{-2}{*}{eng}                       &  \multirow{-2}{*}{ind}               &       \multirow{-2}{*}{jpn}           &   \multirow{-2}{*}{kor}                                        &         \multirow{-2}{*}{tha}         &         \multirow{-2}{*}{tur}        &       \multirow{-2}{*}{vie}         &      \multirow{-2}{*}{\textbf{Avg.}}               \\
\midrule
  \multicolumn{11}{c}{\textbf{Cascaded ASR+MT Models}} \\
\midrule
Whisper-Large-V3 + NLLB-200-3.3B~\cite{nllb2024scaling}      &  \underline{78.3}&\underline{79.8}&83.9&\underline{81.9}&\underline{79.8}&\underline{81.4}&\underline{78.2}&\underline{82.4}&\underline{78.6}&\underline{80.5}      \\   
Whisper-Large-V3 + LLaMAX3-8B-Alpaca~\cite{lu2024llamax}     &   75.3&77.8&80.8&78.8&77.5&78.4&75.6&78.2&76.3&77.6    \\          
\midrule
\multicolumn{11}{c}{\textbf{End-to-end S2TT Models}} \\
\midrule
SeamlessM4T-V2-Large~\cite{seamless2025joint}    & 70.2&73.9&\underline{85.2}&71.7&69.4&74.0&69.1&72.2&72.9&73.2       \\
MCAT-Large-27B (ours)   & \colorbox{cvprblue!14}{\textbf{79.0}}&\colorbox{cvprblue!14}{\textbf{80.6}}&\colorbox{cvprblue!14}{\textbf{86.5}}&\colorbox{cvprblue!14}{\textbf{83.5}}&\colorbox{cvprblue!14}{\textbf{80.0}}&\colorbox{cvprblue!14}{\textbf{82.6}}&\colorbox{cvprblue!14}{\textbf{78.2}}&\colorbox{cvprblue!14}{\textbf{83.2}}&\colorbox{cvprblue!14}{\textbf{79.9}}&\colorbox{cvprblue!14}{\textbf{81.5}}       \\
\bottomrule
\end{tabular}}
\raggedright{\hspace*{1em}\underline{Underlined} denotes previous state-of-the-art models, while \colorbox{cvprblue!14}{highlighted} entries surpass the previous models.}
\label{0970}
\end{table*}
\endgroup

\section{Experiments}
\subsection{Many-to-Many S2TT on FLEURS}
\subsubsection{\textbf{Language Direction}}
To rigorously evaluate the translation performance, we identified $\mathbf{9}$ \textbf{language families} commonly supported by both model variants. Then, we selected the language with the largest speaker population from each family. This methodology yielded two evaluation sets based on translation directions: $\mathbf{9 \times 27}$ set and $\mathbf{9 \times 69}$ set for the MCAT-Small and MCAT-Large models, respectively.

\subsubsection{\textbf{English-centric vs. Balanced Optimization}}
Table \ref{0970} shows that SeamlessM4T-V2-Large performs strongly in English directions (e.g., $85.2$ for $9 \times 69$), suggesting an English-centric design that degrades performance for other pairs (average $73.2$). Conversely, MCAT-Large achieves a better balance, maintaining high English scores ($\mathbf{86.5}$) while significantly boosting non-English performance (average $\mathbf{81.5}$).

\subsubsection{\textbf{Comparative Analysis of ASR and S2TT Tasks}}
Table~\ref{tab:performance_comparison} shows a strong correlation between ASR accuracy and S2TT performance: lower error rates (WER/CER) consistently yield higher COMET scores. 
While English achieves the best results (14.7 WER, 85.9 COMET), a significant gap remains for other languages, suggesting substantial room for improvement in non-English ASR pre-training.

\begin{table}[h]
    \centering
    \small 
        \caption{Comparison of MCAT-Small-9B on S2TT and ASR tasks.}
    \resizebox{1.0\linewidth}{!}{
    \setlength{\tabcolsep}{2pt} 
    \begin{tabular}{lcccccccccc}
        \toprule\\
        \multirow{-2}{*}{{X $\rightarrow$ 27}} & \multirow{-2}{*}{ara} & \multirow{-2}{*}{cmn} & \multirow{-2}{*}{eng} & \multirow{-2}{*}{ind} & \multirow{-2}{*}{jpn} & \multirow{-2}{*}{kor} & \multirow{-2}{*}{tha} & \multirow{-2}{*}{tur} & \multirow{-2}{*}{vie} \\ 
        \midrule
        COMET & 77.8 & {79.8} & \textbf{{85.9}} & {82.3} & 79.7 & {81.9} & {{78.6}} & 82.1 & {79.3}  \\
        WER / CER & 45.2 & {32.3} & \textbf{{14.7}} & {22.5} & 29.6 & {19.5} & {{32.0}} & 32.2 & {29.6}  \\
        \bottomrule
    \end{tabular}}
 \raggedright{\hspace*{0.5em}  CER is used for cmn, jpn, kor, and tha, while WER is for others.}
    \label{tab:performance_comparison}
\end{table}

\begin{figure}[t]
\centering
\includegraphics[width=\linewidth]{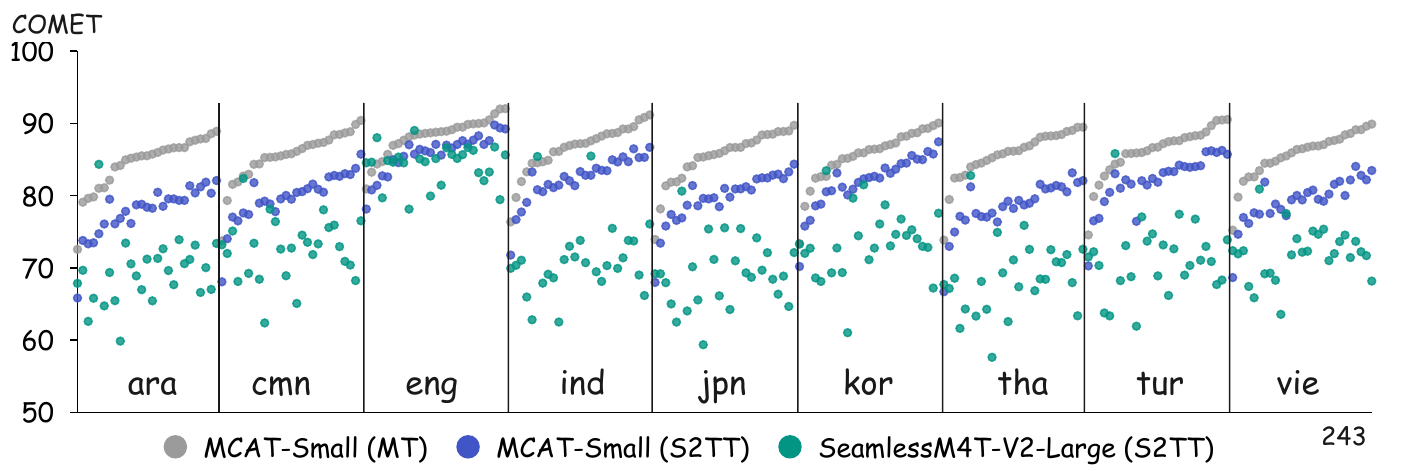}
\caption{\textbf{COMET Scores Between MT and S2TT on 9x27 directions.} The results show a
strong correlation, suggesting that our S2TT capability
is derived from the MT model.}
\label{mt_s2tt}
\end{figure}

\subsubsection{\textbf{MT vs. S2TT}}
As shown in Figure \ref{mt_s2tt}, since MCAT-Small-9B is built upon an LLM backbone, it inherently retains MT capabilities. We compared MT and S2TT performance across $9 \times 27$ directions. The results reveal a strong correlation between the MLLM's S2TT and MT performance, confirming that its robust S2TT capability primarily stems from the underlying LLM's machine translation proficiency.

\subsubsection{\textbf{Cascaded Systems vs. End-to-End Models}}
As shown in Table \ref{0970}, previous end-to-end models excelled primarily in English directions due to abundant English-aligned $\text{S2TT}$ data, but often underperformed compared to cascaded systems in non-English directions. Our method establishes a new comprehensive state-of-the-art across all nine representative source language directions.

\subsubsection{\textbf{Scaling Law of Language Coverage}}
MLLMs typically suffer from catastrophic forgetting during language expansion. However, Table \ref{0970} shows that MCAT-Large-27B maintains consistent performance, averaging $\mathbf{81.3}$ and $\mathbf{81.5}$ across 27 and 69 languages, respectively. This stability demonstrates that our data balancing and training strategies effectively mitigate performance degradation, ensuring robust language scalability.

\newpage

\begingroup
\renewcommand{\arraystretch}{1.3} 
        \setlength{\tabcolsep}{5pt} 
\begin{table*}[!ht]
\centering
\small
\caption{COMET Results on eng $\rightarrow$ 27 Directions on the FLEURS Dataset.}
\resizebox{1.0\textwidth}{!}{
\begin{tabular}{l|ccccccccccccccccccccccc}
\toprule
\rowcolor{gray!12}
  &                     &                    &                    &                   &                    &      &                   &                     &                    &           &       & &  &                    \\
\rowcolor{gray!12}
\multirow{-2}{*}{\textbf{End-to-end Models} (eng $\rightarrow$ X)}                                                      &   \multirow{-2}{*}{ara}              &    \multirow{-2}{*}{ben}              &    \multirow{-2}{*}{ces}                       &  \multirow{-2}{*}{cmn}               &       \multirow{-2}{*}{deu}           &   \multirow{-2}{*}{fas}                        &         \multirow{-2}{*}{fra}         &         \multirow{-2}{*}{heb}        &       \multirow{-2}{*}{hin}               &   \multirow{-2}{*}{ind}                        &         \multirow{-2}{*}{ita}         &         \multirow{-2}{*}{jpn}              &         \multirow{-2}{*}{khm}  &         \multirow{-2}{*}{kor}  \\
\midrule
SeamlessM4T-V2-Large~\cite{seamless2025joint} &    84.5&\underline{84.6}&88.0&79.7&84.8&84.6&85.3&\underline{84.5}&\underline{78.1}&89.0&85.1&84.7&\underline{79.9}&85.1         \\
ZeroSwot-Large~\cite{tsiamas-etal-2024-pushing} & 83.1 & 82.1 & 76.1 & 81.0 & 82.4 & 83.4 & 82.5 & 75.8 & 75.3 & 86.8 & 81.3 & 86.2 & 75.7 & 75.5\\
Step-Audio-2-mini~\cite{wu2025step} & 83.8 & 49.5 & 79.3 & 87.2 & 83.4 & 53.2 & 84.7 & 53.2 & 46.8 & 82.3 & 84.1 & 81.3 & 33.3 & 82.5\\
Qwen2.5-Omni-7B~\cite{Qwen2.5-Omni}     & 84.5&63.0&81.9&86.5&85.0&73.4&84.6&66.1&58.7&86.2&84.2&88.3&37.8&84.7 \\
Qwen3-Omni-30B-A3B-Instruct~\cite{xu2025qwen3}   & \underline{86.6}&83.3&\underline{89.0}&\underline{88.3}&\underline{86.8}&\underline{85.5}&\underline{87.3}&73.9&77.7&\underline{90.9}&\underline{87.3}&\underline{90.8}&77.6&\underline{89.7} \\
MCAT-Small-9B (ours)    &  85.4&\colorbox{cvprblue!14}{85.0}&\colorbox{cvprblue!14}{89.4}&87.0&85.7&\colorbox{cvprblue!14}{86.4}&85.6&\colorbox{cvprblue!14}{86.6}&\colorbox{cvprblue!14}{78.1}&89.2&86.0&89.8&\colorbox{cvprblue!14}{\textbf{81.4}}&88.3\\
MCAT-Large-27B (ours) &86.1&\colorbox{cvprblue!14}{\textbf{85.2}}&\colorbox{cvprblue!14}{\textbf{90.1}}&87.2&86.5&\colorbox{cvprblue!14}{\textbf{86.9}}&86.3&\colorbox{cvprblue!14}{\textbf{87.2}}&\colorbox{cvprblue!14}{\textbf{78.9}}&90.2&86.8&90.4&79.7&88.5\\ \midrule
\rowcolor{gray!12}
  &                     &                    &                    &                   &                    &      &                   &                     &                    &      &                     &        &             &      \multicolumn{1}{|c}{}                              \\
\rowcolor{gray!12}
\multirow{-2}{*}{\textbf{End-to-end Models} (eng $\rightarrow$ X)}                                                      &   \multirow{-2}{*}{lao}              &    \multirow{-2}{*}{msa}   &    \multirow{-2}{*}{mya}             &    \multirow{-2}{*}{nld}                       &  \multirow{-2}{*}{pol}               &       \multirow{-2}{*}{por}           &   \multirow{-2}{*}{rus}                        &         \multirow{-2}{*}{spa}         &         \multirow{-2}{*}{tgl}        &       \multirow{-2}{*}{tha}               &   \multirow{-2}{*}{tur}                        &         \multirow{-2}{*}{urd}         &         \multirow{-2}{*}{vie}    &         \multicolumn{1}{|c}{\multirow{-2}{*}{\textbf{Avg.}}}                      \\
        
\midrule
SeamlessM4T-V2-Large~\cite{seamless2025joint} &  \underline{81.4}&86.6&\underline{85.7}&85.1&85.7&86.6&86.3&83.2&\underline{82.1}&83.3&86.7&\underline{79.4}&85.6&\multicolumn{1}{|c}{84.3}\\
ZeroSwot-Large~\cite{tsiamas-etal-2024-pushing} & 76.8 & 83.6 & 83.4 & 82.1 & 80.4 & 80.0 & 83.2 & 74.5 & 77.0 & 73.2 & 84.4 & 78.0 & 82.8 & \multicolumn{1}{|c}{80.2}\\
Step-Audio-2-mini~\cite{wu2025step} & 32.4 & 77.4 & 32.3 & 77.4 & 71.3 & 87.1 & 84.1 & 83.8 & 53.5 & 66.9 & 66.4 & 38.3 & 75.7 & \multicolumn{1}{|c}{67.8}\\
Qwen2.5-Omni-7B~\cite{Qwen2.5-Omni}    &  38.7&83.5&41.1&82.1&81.5&86.7&85.2&83.4&55.6&82.1&79.2&51.6&76.4&\multicolumn{1}{|c}{73.8}\\
Qwen3-Omni-30B-A3B-Instruct~\cite{xu2025qwen3}  &80.2&\underline{88.3}&71.9&\underline{86.2}&\underline{87.3}&\underline{88.4}&\underline{88.9}&\underline{85.4}&80.1&\underline{88.8}&\underline{88.1}&78.3&\underline{88.5}&\multicolumn{1}{|c}{\underline{85.0}}  \\
MCAT-Small-9B (ours)    &   \colorbox{cvprblue!14}{82.6}&87.1&\colorbox{cvprblue!14}{\textbf{87.0}}&\colorbox{cvprblue!14}{86.2}&\colorbox{cvprblue!14}{87.6}&87.1&87.7&84.6&\colorbox{cvprblue!14}{82.7}&87.1&87.6&\colorbox{cvprblue!14}{80.8}&87.1&\multicolumn{1}{|c}{\colorbox{cvprblue!14}{85.9}}\\
MCAT-Large-27B (ours) &    \colorbox{cvprblue!14}{\textbf{83.1}}&87.5&85.0&\colorbox{cvprblue!14}{\textbf{86.6}}&\colorbox{cvprblue!14}{\textbf{88.1}}&87.8&88.8&85.2&\colorbox{cvprblue!14}{\textbf{83.3}}&87.7&\colorbox{cvprblue!14}{\textbf{88.2}}&\colorbox{cvprblue!14}{\textbf{80.9}}&87.8&\multicolumn{1}{|c}{\colorbox{cvprblue!14}{\textbf{86.3}}}  \\
\bottomrule
\end{tabular}}
\raggedright{\hspace*{1em}\underline{Underlined} denotes previous state-of-the-art models, while \colorbox{cvprblue!14}{highlighted} entries surpass the previous models.}
\label{0128}
\end{table*}
\endgroup

\begin{figure*}[t]
\centering
\includegraphics[width=\linewidth]{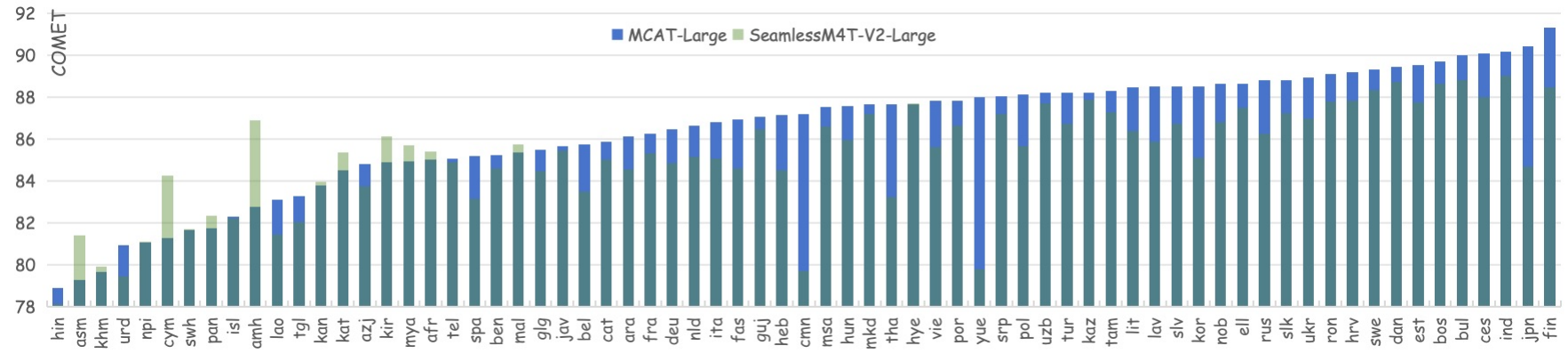} 
\caption{\textbf{COMET Scores for the English$\to$69 Translation Directions on the FLEURS Dataset.} The blue bars denote stronger translation performance for the MCAT-Large model in a total of 55 directions.}
\label{0169}
\end{figure*}

\newpage

\subsection{Eng$\to$X S2TT on FLEURS}
Table \ref{0128} presents a comprehensive comparison of the performance of various end-to-end translation models across 27 target language directions originating from English, evaluated using the COMET metric. Our proposed models, MCAT-Small and MCAT-Large, show competitive and often superior results compared to established models like SeamlessM4T-V2-Large and Qwen2.5-Omni-7B. Specifically, \textbf{MCAT-Large} achieves the highest overall average COMET score of $\mathbf{86.3}$, surpassing the best prior model, $\text{Qwen3-Omni-30B-A3B-Instruct}$.

\subsubsection{\textbf{Comparison on Eng$\to$27 Language Directions}}
As shown in Table \ref{0128}, we compared the performance of end-to-end models on English. It can be observed that the models in the Qwen-Omni series show strong performance on high-resource languages such as $\mathbf{cmn}$ and $\mathbf{fra}$, but exhibit noticeable deficiencies in low-resource languages like $\mathbf{khm}$ and $\mathbf{mya}$. In contrast, our model achieves competitive performance on high-resource languages while demonstrating powerful performance on low-resource languages.

\subsubsection{\textbf{Comparison on Eng$\to$69 Directions}}
As Figure \ref{0169} illustrates, MCAT-Large shows a consistent performance advantage over the \textbf{SeamlessM4T-V2-Large} baseline, particularly in mid-to-high-resource settings (e.g., $\mathbf{tgl}$, $\mathbf{cmn}$). Quantitatively, MCAT-Large achieved superior results in $\mathbf{55}$ out of $69$ tested directions, confirming a clear overall edge. However, its relatively weaker performance on low-resource languages (e.g., $\mathbf{amh}$, $\mathbf{cym}$) is primarily constrained by the intrinsic capabilities of the underlying LLM component within the MLLM architecture.
\subsubsection{\textbf{MCAT-Small-9B vs. Qwen2.5-Omni-7B}}
As shown in Table \ref{0128}, \textbf{MCAT-Small} consistently surpasses \textbf{Qwen2.5-Omni-7B} across all 27 translation directions, achieving a substantial average COMET gain of $\mathbf{12.1}$ points (from 73.8 to 85.9). Notably, its overall performance is comparable to that of Qwen3-Omni-30B-A3B-Instruct, underscoring the efficiency of our architecture. These results demonstrate that \textbf{MCAT-Small} provides competitive translation quality while maintaining lower computational and resource requirements.

\newpage
\begin{figure*}[t]
\centering
\includegraphics[width=\linewidth]{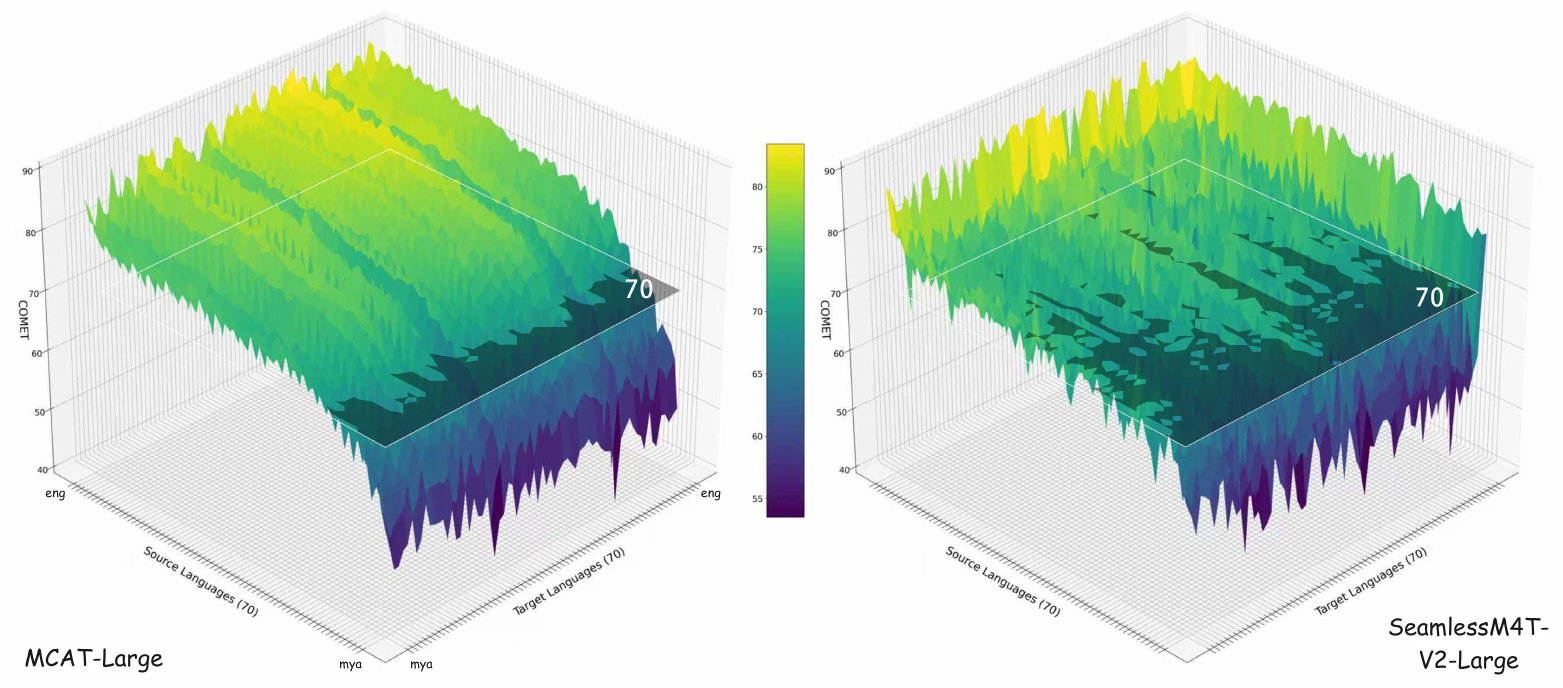}
\caption{\textbf{COMET Scores Across $70\times70$ Translation Directions.} For cases like $\text{eng}\to\text{eng}$, no score is calculated, and smoothing was applied in the figure.}
\label{7070}
\end{figure*}

\begingroup
\renewcommand{\arraystretch}{1.2} 
        \setlength{\tabcolsep}{5pt} 
\begin{table}[t]
\centering
\small
\caption{COMET Scores Statistics on the FLEURS Dataset.}
\resizebox{\linewidth}{!}{
\begin{tabular}{l|cccc|c}
\toprule

  &                     &                    &                    &                   &                                                    \\

\multirow{-2}{*}{\textbf{Models}}                                             &       \multirow{-2}{*}{$x\ge$90}          &   \multirow{-2}{*}{90$>x\ge$80}              &    \multirow{-2}{*}{80$>x\ge$70}              &    \multirow{-2}{*}{$x<$70}                                       &       \multirow{-2}{*}{Total}           \\
\midrule
MCAT-Small-9B &0&399&265&92&$28\times27$ \\
MCAT-Large-27B &6 &2197&1834& 793 &$70\times69$\\
SeamlessM4T-V2-Large &0 &215&2719& 1896&$70\times69$\\
\bottomrule
\end{tabular}}
\label{7069}
\end{table}
\endgroup

\subsection{COMET Score Across 70 Languages}
\subsubsection{\textbf{Comparision on 70 Languages}}
As shown in Figure \ref{7070}, the surface is predominantly colored yellow and light green, corresponding to $\text{COMET}$ scores well above 70. This signifies that the model provides usable translations across the vast majority of the potential language pairs.
Furthermore, Figure \ref{70_avg} shows the translation performance of S2TT for 70 languages, ordered from smallest to largest average performance.
\subsubsection{\textbf{Multilingual Consistency}}
As shown in Figure \ref{7070}, the MCAT model demonstrates a strong degree of multilingual consistency across translation directions. Specifically, for any given source language, the COMET scores when translating into the wide range of target languages are observed to be relatively uniform and cluster within a tight range. This consistency is a critical indicator of the model's design success. It strongly suggests that the model is successfully employing shared knowledge components and parameter sharing across its multilingual capacity.
\subsubsection{\textbf{Quantitative Confirmation of Robustness}}
Table \ref{7069} provides a quantitative distribution of the COMET scores into specific score bins for two models. For the $\text{MCAT-Large-27B}$ model, the majority of the $70 \times 69$ directions ($4,830$ pairs) fall into the high-score brackets. Specifically, $4,037$ pairs (combining the $70-80$, $80-90$, and $>90$ bins) achieve a $\text{COMET}$ score exceeding $70$. 

 \begin{figure}[t]
 \centering 
    \includegraphics[width=\linewidth]{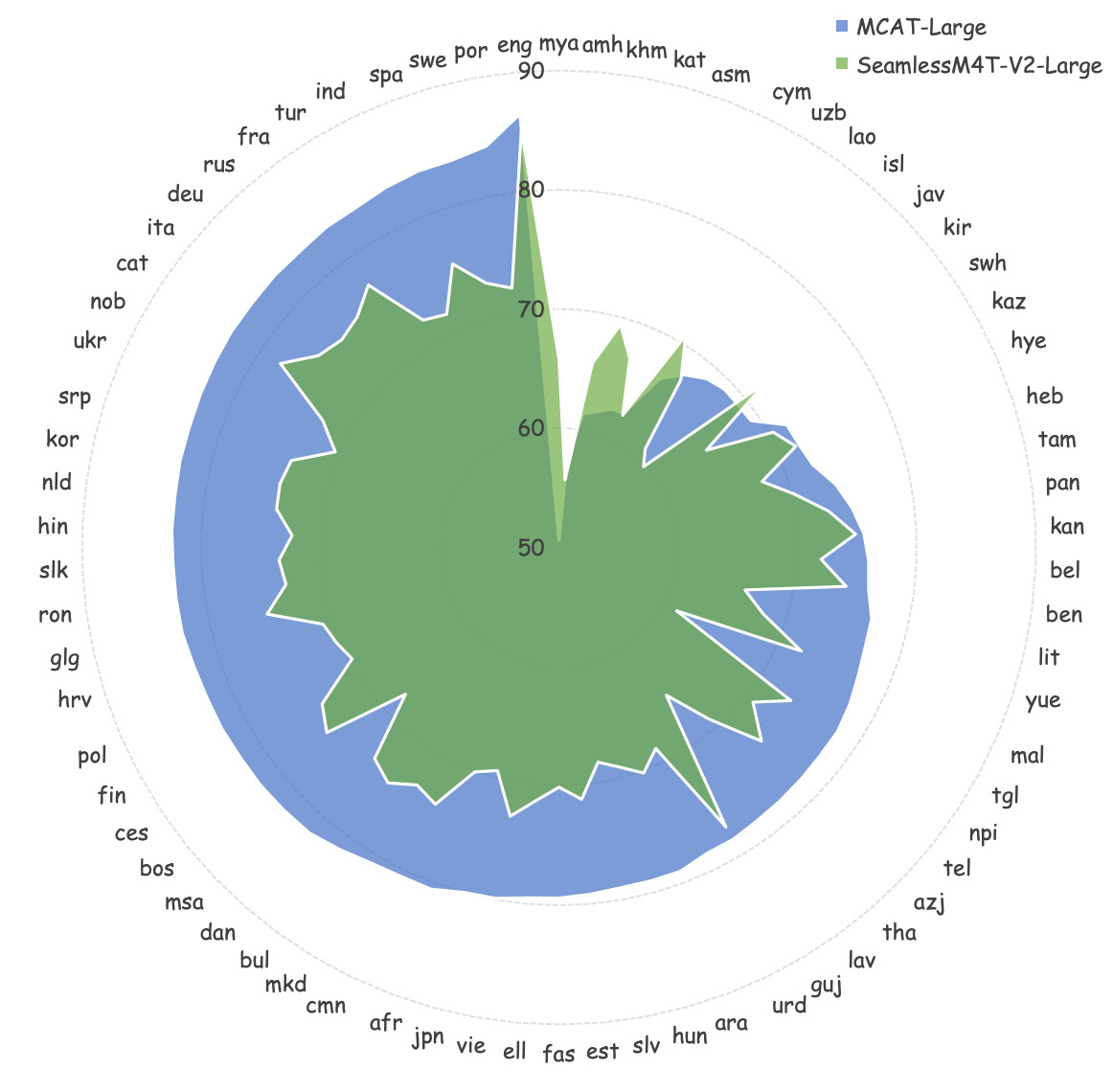}
    \caption{\textbf{Average Performance Across 70 Languages.}}
    \label{70_avg}
\end{figure}

\subsubsection{\textbf{Asymmetry in Low-Resource Language}}
As shown in Figure \ref{7070}, for languages such as $\text{\textbf{mya}}$ (Burmese), $\text{\textbf{amh}}$ (Amharic), and $\text{\textbf{khm}}$ (Khmer), the $\text{COMET}$ scores are very \textbf{high} when these languages serve as the \textbf{target language}. However, their scores are extremely \textbf{low} when they act as the \textbf{source language}, as shown in Figure \ref{70_avg}. This suggests that the MLLM possesses sufficient capability to understand and generate text in these languages; however, the scarcity of speech recognition data prevents accurate speech decoding, leading to low overall scores. This finding implicitly suggests a critical need for more ASR data for these specific languages.

\newpage

\begin{table*}[t]
  \centering
    \caption{Comet Result of Ablation Studies on the FLEURS Dataset. 
}
  \renewcommand{\arraystretch}{1.2}  
        \setlength{\tabcolsep}{6pt} 

    \resizebox{1.0\textwidth}{!}{
  \begin{tabular}{lcccccccccccccccc} \toprule 
  \multirow{2}{*}{COMET eng $\rightarrow$ X}& \multirow{2}{*}{ara}& \multirow{2}{*}{cmn}& \multirow{2}{*}{ind} &\multirow{2}{*}{jpn}&\multirow{2}{*}{khm}&\multirow{2}{*}{kor}&\multirow{2}{*}{lao}&\multirow{2}{*}{mya}&\multirow{2}{*}{tha}&\multirow{2}{*}{tur}&\multirow{2}{*}{vie}&\multirow{2}{*}{\textbf{Avg.}}\\
    & \\ \midrule

          MCAT-Small-9B & 85.4&87.0&89.2&89.9&81.5&88.2&82.5&87.1&87.1&87.5&87.2&86.6 \\
                  \hspace{1em}w/o ASR  & 43.6&49.2&47.7&51.9&44.3&45.1&49.9&51.3&47.2&49.9&60.0&49.1 {\color{red}(-37.5)} \\
                  \hspace{1em}w/o SMT  & 85.2&86.9&89.1&89.7&81.2&88.0&82.2&87.0&87.0&87.4&86.8&86.4 {\color{red}(-0.2)} \\
                              \hspace{1em}w/o SRT  & 75.3&76.1&80.7&79.6&74.0&76.7&73.8&78.4&76.9&75.5&76.6&76.7 {\color{red}(-9.9)} \\
                              \hspace{1em}w/o Balance & 
                              85.1 & 86.2 &88.7 &89.3 &80.9 &87.8 &82.1 &86.5 &86.3 &87.3 &86.7 &86.1 {\color{red}(-0.5)} 
                              \\
         \hspace{1em}w/o LLM Lora & 85.6&86.9&89.3&89.6&81.3&88.1&82.6&87.0&87.0&87.6&87.1&86.5 {\color{red}(-0.1)} \\
         \hspace{1em}+ Beam Search 5 & 86.3 &87.5 &89.9 &90.3 &83.7 &88.6 &84.8 &88.5 &87.9 &88.4 &87.7 &87.6  {\color{acl_green}(+1.0)}\\

\bottomrule
  \end{tabular}}

  \label{tab:ablation}
\end{table*}

\subsection{Ablation Study}

\subsubsection{\textbf{Curriculum Learning Strategy}} 
The ablation study in Table \ref{tab:ablation} validates our three-stage curriculum learning strategy. 
First, \textbf{ASR pre-training} is the foundational pillar; its removal (\textit{w/o ASR}) causes a catastrophic COMET score drop from 86.6 to 49.1, confirming its role in establishing robust speech representations. 
Second, \textbf{SMT enhancement} provides critical refinement and accelerates convergence; omitting it (\textit{w/o SMT}) results in a slight score decrease to 86.4 and a more protracted training process. 
Third, \textbf{SRT activation} is essential for many-to-many translation; without it (\textit{w/o SRT}), performance declines significantly to 76.7, highlighting its role in activating the LLM's cross-lingual reasoning. 
Overall, this curriculum serves as a vital scaffold for knowledge acquisition, particularly when direct instruction tuning is insufficient in low-data scenarios.
\subsubsection{\textbf{Data Balancing Strategy}}
The impact of our data balancing strategy is evidenced by the comparison in Table \ref{tab:ablation}. Removing this component (\textit{w/o Data Balance}) leads to a performance decline of 0.5 points in the average COMET score (from 86.6 to 86.1). 
This regression suggests that balanced sampling is crucial for mitigating the bias toward high-resource languages during many-to-many training. By ensuring equitable exposure to diverse language pairs, the strategy prevents the model from over-fitting to dominant directions.

\subsubsection{\textbf{Adapter Training vs. LLM Fine-tuning}}
As shown in Table \ref{tab:ablation}, evaluating LoRA-based fine-tuning reveals that training only the speech encoder and projector (\textit{w/o LLM LoRA}) yields surprisingly strong performance. This suggests that pre-trained LLMs possess inherent cross-lingual reasoning capabilities that require only minimal speech-feature alignment. However, incorporating LoRA (\textit{w/ LLM LoRA}) provides marginal yet consistent gains across all language pairs.

\subsubsection{\textbf{Beam Search vs. Greedy Search}}
We further evaluate the impact of decoding strategies by comparing Beam Search with Greedy Search in Table \ref{tab:ablation}. 
Switching from Greedy Search to Beam Search with a width of 5 (\textit{+ Beam Search 5}) yields a significant performance gain of 1.0 average COMET points, raising the score from 86.6 to 87.6. 
 The results confirm that Beam Search is essential for maximizing the linguistic fluency and translation accuracy of the MCAT model.

\newpage

\subsection{Discussion}

\subsubsection{\textbf{Model Architecture}}
We draw inspiration from the design in~\cite{gaido-etal-2024-speech,tang2024salmonn}.
\textbf{For the speech encoder}: Since the model is designed for S2TT tasks, we aim to support as many languages as possible. The Whisper encoder is an excellent choice as it has proven its effectiveness across multiple languages; however, its primary limitation is that it can only process audio segments up to 30 seconds in length. 

\textbf{For the speech adapter}: Since we employ the Whisper encoder, the input Mel spectrogram is padded to a length of 3000 and yields an encoded output of 1500. Our design prioritizes maximal sequence compression. To achieve this, we utilize a Q-Former to condense the output to 150 tokens, then apply average pooling to further reduce the sequence to 30 tokens.

\textbf{For the LLM}: We aim for the LLM to support a wide range of languages while maintaining robust foundational machine translation capabilities. Our evaluations demonstrated that the Gemma series consistently delivers superior performance; consequently, we selected GemmaX2 and Gemma-3-27B as our base models.

\subsubsection{\textbf{Scaling Law of Data}}
Table \ref{tab:covost2} validates the data scaling law by demonstrating that performance is strongly bounded by training data volume even with a fixed architecture. Expanding the English training data for the MCAT-Small-9B model from the low-data FLEURS regime (\textbf{7.5 h}) to the larger CoVoST-2 dataset (\textbf{429.6 h}) yields a dramatic performance surge. This $\sim$57$\times$ increase in data volume boosts the average COMET score from 81.7 to \textbf{85.6}. These results confirm that scaling the data corpus remains an essential driver for improving translation quality in speech-to-text tasks.

 \begin{table}[h]
  \centering
    \small
      \caption{Scaling law of Data.
}
        \renewcommand{\arraystretch}{1.2} 
          \setlength{\tabcolsep}{3pt} 
              \resizebox{1.0\linewidth}{!}{
  \begin{tabular}{lcccccccccc} \toprule 
    \multirow{2}{*}{COMET  eng$\rightarrow$X}&   \multicolumn{8}{c}{CoVoST-2}\\ 
      & ara&cmn &deu & fas&ind&jpn&tur &\textbf{Avg.} \\ \midrule    
      \multicolumn{8}{l}{\textbf{with FLEURS eng data: 7.5 h}}\\ 
      {MCAT-Small-9B} &79.8&82.0&79.7&80.3&83.7&85.1&81.3&81.7\\
      \multicolumn{8}{l}{\textbf{with CoVoST-2 eng data: 429.6 h}} \\ 
         {MCAT-Small-9B-V2} & 83.8&86.0&84.3&83.8&88.2&87.9&85.4&85.6\\ \bottomrule

  \end{tabular}}
  \label{tab:covost2}
\end{table}

\newpage

\subsubsection{\textbf{Inference Speed}}As shown in Table \ref{tab:speed}, our \textbf{MCAT models} demonstrate superior computational efficiency compared to Qwen2.5-Omni- 7B under the same BF16 A100 GPU setup. This advantage stems from our optimized architecture which utilizes only \textbf{30 tokens per speech sample} compared to the 750 tokens typically required by Qwen2.5-Omni. Specifically, the {MCAT-Small-9B} model achieves a remarkable inference time of only \textbf{96 seconds} using the vLLM~\cite{kwon2023efficient} framework with dynamic batching. This represents a $\mathbf{3.37\times}$ speedup over the 323 seconds required by Qwen2.5-Omni-7B. Even the significantly larger {MCAT-Large-27B} model remains substantially faster than the baseline, completing the same task in 182 seconds and further validating the efficiency of our sequence compression strategy.

\begin{table}[h]
  \centering
  \caption{Inference performance comparison (Single A100 GPU).}
  \label{tab:speed}
  \renewcommand{\arraystretch}{1.1}
  \setlength{\tabcolsep}{6pt} 
  \resizebox{1.0\linewidth}{!}{
  \begin{tabular}{lcccc} \toprule 
  &&&&\\
    \multirow{-2}{*}{Model} & \multirow{-2}{*}{Framework} & \multirow{-2}{*}{Beam} & \multirow{-2}{*}{Batch} & \multirow{-2}{*}{Time (s)} \\ \midrule
    Qwen2.5-Omni-7B & vLLM & 1 & Dynamic & 323 \\ \midrule
    
    \multirow{3}{*}{MCAT-Small-9B} & \multirow{2}{*}{Transformer} & 1 & 50 & 154 \\
     & & 5 & 50 & 226 \\ \cmidrule(lr){2-5}
     & vLLM & 1 & Dynamic & 96 \\ \midrule

    \multirow{3}{*}{MCAT-Large-27B} & \multirow{2}{*}{Transformer} & 1 & 50 & 304 \\
     & & 5 & 50 & 415 \\ \cmidrule(lr){2-5}
     & vLLM & 1 & Dynamic & 182 \\ 
    \bottomrule
  \end{tabular}}
  \raggedright{\hspace*{0.5em} Results are measured on 1,000 speech samples using BF16 precision.}
\end{table}

\section{Conclusion}
We successfully addressed the critical language scalability and efficiency constraints of MLLMs for the S2TT task. Our primary contributions are twofold: we introduced a novel multilingual S2TT training strategy leveraging curriculum learning for mutual translation across $\mathbf{70}$ languages ($\mathbf{4,830}$ directions), and we designed an efficient architecture with an optimized speech adapter that achieved a $\mathbf{25\times}$ input compression (reducing tokens from 750 to 30). Crucially, our models ($\mathbf{9B}/\mathbf{27B}$) surpassed state-of-the-art end-to-end performance on the FLEURS dataset across $\mathbf{70 \times 69}$ directions, despite the extreme compression. This high performance and extensive multilingual support is attained with remarkable resource efficiency, requiring only $\mathbf{\sim 100M}$ trainable parameters and limited data resources ($10$h S2TT data per language).

\section*{Limitations}
This paper presents a method for training an MLLM for languages with less than 10 hours of speech translation data.

However, the performance of S2TT and the range of supported languages are constrained by the capabilities of the LLM. MLLMs trained using this method may not perform well on languages that are not supported by the LLM or on those with poor machine translation performance. Furthermore, for some low-resource languages, additional speech recognition data is still required for initialization.

\bibliographystyle{IEEEtran}
\bibliography{custom} 

\newpage
\appendix
\subsection{Language Coverage}
The MLLM's S2TT capability is contingent upon the upper bound of the underlying LLM's MT performance. Consequently, the MT capability of the base model directly determines the ceiling of our translation quality and guides our final selection of supported languages.

\subsubsection{\textbf{28 Languages for MCAT-Small}}
The GemmaX2-9B~\cite{cui2025multilingual} model was specifically trained and optimized for these 28 target languages, resulting in a significant performance improvement. Based on this design, these 28 languages are designated as fully supported.

\subsubsection{\textbf{70 Languages for MCAT-Large}}
To determine the scope of supported languages, we evaluated the translation quality of the Gemma3-27B \cite{team2025gemma} base model using the Flores \cite{nllb2024scaling} dataset. Based on {empirical observations}, an {empirical threshold} of 70 for the COMET score was set as the {acceptable requirement} for language selection. Approximately 70 languages reached or exceeded this benchmark, leading to their inclusion in the final support range for MCAT-Large.

\subsection{Evaluation Metrics: COMET vs. spBLEU}
As shown in Table \ref{0970} and \ref{0970_spbleu}, a notably divergent trend is observed for the average scores in the 9$\to$69 direction between the cascaded NLLB model ($21.0 / 80.5$) and MCAT-Large ($20.1 / 81.5$). Specifically, NLLB achieves a higher spBLEU score but a lower COMET score compared to MCAT-Large. This phenomenon is rooted in the distinct design philosophies of the models and the metrics: NLLB, as a specialized machine translation model, is optimized for strict sentence-level alignment and high lexical overlap with the reference translations, leading to superior performance on the n-gram-based spBLEU~\cite{post-2018-call}. 
In contrast, MCAT-Large, an MLLM-based architecture, prioritizes generating fluent and human-natural sentences through flexible paraphrasing and semantic preservation. This semantic quality and fluency, which may come at the expense of rigid word-for-word matching, is better captured by COMET~\cite{rei2022comet}, a neural metric that has demonstrated a higher correlation with human judgment of translation quality.

\subsection{Data Duration Across 70 Languages on Fleurs}
 As  shown in Table \ref{time}, the Fleurs dataset comprises segments with an average duration of 12.3s. The majority of samples (approximately 61\%) fall within the 10–30s range, while segments shorter than 5s account for only a small fraction (679 samples).
\begingroup
\renewcommand{\arraystretch}{1.3} 
\setlength{\tabcolsep}{8pt}      
\begin{table}[h]
\centering
\small
\caption{Data Time Across 70 Languages.}
\label{time}
\resizebox{\linewidth}{!}{
\begin{tabular}{l|cccc|c}
\toprule
\textbf{Duration} & $x < 5$ & $5 \le x < 10$ & $10 \le x < 15$ & $15 \le x < 30$ & \textbf{Avg Time} \\
\midrule
Fleurs & 679 & 17,885 & 22,454 & 12,326 & 12.3 \\
\bottomrule
\end{tabular}}
\end{table}
\endgroup

\begin{table*}[h]
\centering
\small
\caption{Summary of Training Datasets for MCAT Model on 8 A100 GPUs. }
\resizebox{\textwidth}{!}{%
\begin{tabular}{lccccccc}
\toprule
\textbf{Model} & \textbf{Task} & \textbf{Dataset} & \textbf{Split}& \textbf{Data Size}&\textbf{Batch} &\textbf{Step}& \textbf{Metric} \\
\midrule
\multirow{7}{*}{MCAT-Large} & \multirow{2}{*}{ASR} & Common Voice 22 & train &$\sim$3500h& \multirow{2}{*}{8}  & \multirow{1}{*}{180000} & \multirow{2}{*}{WER $\downarrow$} \\
& & FLEURS & train &$\sim$617.7h& & \multirow{1}{*}{54000}\\
\cmidrule(lr){2-8}
& \multirow{2}{*}{SMT} & \multirow{2}{*}{FLEURS} & \multirow{2}{*}{train} &\multirow{2}{*}{$\sim$617.7h} & \multirow{2}{*}{3}  & \multirow{2}{*}{16000} & \multirow{2}{*}{spBLEU / COMET $\uparrow$} \\
& & & & &&  \\ \cmidrule(lr){2-8}
& \multirow{2}{*}{SRT}  & \multirow{2}{*}{FLEURS} & \multirow{2}{*}{train} &\multirow{2}{*}{$\sim$617.7h} & \multirow{2}{*}{3}  & \multirow{2}{*}{96000}& \multirow{2}{*}{spBLEU / COMET $\uparrow$} \\
& & & & &&  \\
\bottomrule
\end{tabular}}
\raggedright{Data size refers to the actual amount used, as we removed overly long samples and balanced the data across different languages.}
\label{tab:dataset}
\end{table*}

\begin{CJK*}{UTF8}{gbsn}
\begin{table*}[h]  
    \renewcommand{\arraystretch}{1.1} 
         \caption{\textbf{Case Study.} We compare the BLEU scores of our models with SeamlessM4T-V2.}
    \centering
    \setlength{\tabcolsep}{4pt} 
    \setlength{\extrarowheight}{1pt}
    \resizebox{1\textwidth}{!}{
    \begin{tabular}{llc}
    \toprule
    \textbf{Case} &  & \textbf{COMET}$\uparrow$ \\ \midrule

    \multirow{2}{*}{\textbf{Audio \# 1}} & 它让玩家可以通过在空中移动设备来控制电子游戏中的运动和操作。 &  \\
     & \textcolor{blue}{ This will allow players to control actions and movements in video games by moving the device through the air.} &  \\ \cmidrule(r){1-2}

    \textbf{SeamlessM4T-V2} & \textcolor{purple}{It allows players to control the movement and operation of electronic games through aerial mobile devices.} & 85.5 \\ \cmidrule{2-2}

    \multirow{2}{*}{\textbf{MCAT-Small-9B}} &这让玩家可以通过在空中移动设备来控制电子游戏中的运动和操作。\texttt{<|cmn|><|eng|>} &  \\
    & \textcolor{purple}{This allows the player to control movement and actions in the video game by moving the device in the air.} & 91.1\\ 
    \midrule

    \multirow{2}{*}{\textbf{Audio \# 2}} & In fact, it is not easy to find at all even if one knew it existed. Once inside the cave, it is a total isolation. &  \\
     & \textcolor{blue}{ 事实上，即使知道它的存在，也不容易找到。一旦进入洞穴，就完全与世隔绝了。
} &  \\ \cmidrule(r){1-2}

    \textbf{SeamlessM4T-V2} & \textcolor{purple}{事实上,即使有人知道它存在, 但一旦进入洞穴,} & 69.9 \\ \cmidrule{2-2}

    \multirow{2}{*}{\textbf{MCAT-Small-9B}} & In fact, it is not easy to find at all! Even if one knew it existed, once inside the cave it is a total isolation.\texttt{<|eng|><|cmn|>} &  \\ 
    & \textcolor{purple}{事实上，它根本就不容易找到！即使有人知道它的存在，一旦进入洞穴，就会完全与外界隔绝。} & 92.1  \\ 

    \bottomrule

    \end{tabular}}

    \label{tab:case_study2}
\end{table*}
\end{CJK*}

\begin{table*}[h]
\renewcommand{\arraystretch}{1.1} 
        \setlength{\tabcolsep}{6pt} 
\centering
\small
\caption{spBLEU Results on 9×27 and 9×69 Directions on the FLEURS dataset.  
}
\resizebox{1.0\textwidth}{!}{
\begin{tabular}{l|ccccccccc|c}
\toprule
\rowcolor{gray!12}
  &                     &                    &                    &                   &                    &      &                   &                     &                    &                                        \\
\rowcolor{gray!12}
\multirow{-2}{*}{\textbf{Systems} (X $\rightarrow$ 27)}                                                      &   \multirow{-2}{*}{ara}              &    \multirow{-2}{*}{cmn}              &    \multirow{-2}{*}{eng}                       &  \multirow{-2}{*}{ind}               &       \multirow{-2}{*}{jpn}           &   \multirow{-2}{*}{kor}                        &         \multirow{-2}{*}{tha}         &         \multirow{-2}{*}{tur}        &       \multirow{-2}{*}{vie}    &          \multirow{-2}{*}{\textbf{Avg.}}                    \\
\midrule
\multicolumn{11}{c}{\textbf{Cascaded ASR+MT Models}} \\
\midrule
Whisper-Large-V3 + NLLB-200-3.3B~\cite{nllb2024scaling}&21.5&18.9&30.2&24.2&19.4&19.9&17.5&23.7&18.9&21.6             \\   
Whisper-Large-V3 + LLaMAX3-8B-Alpaca~\cite{lu2024llamax}  &   17.5&16.1&25.2&20.6&16.1&16.9&14.5&19.0&16.2&18.0         \\  
\midrule
\multicolumn{11}{c}{\textbf{End-to-end S2TT Models}} \\
\midrule
SeamlessM4T-V2-Large~\cite{seamless2025joint} &     15.5&13.2&30.9&15.3&12.2&14.4&11.5&15.7&13.9&15.8   \\
MCAT-Small-9B (ours)    &20.2&18.4&32.7&24.5&18.9&21.4&16.8&23.5&18.8&21.7      \\
MCAT-Large-27B (ours) &     20.2&17.9&31.7&24.4&18.3&21.1&15.7&24.2&18.2&21.3    \\
\toprule

\rowcolor{gray!12}
   &                     &                     &                     &                    &                   &   &                   &                    &                   &                           \\
\rowcolor{gray!12}
\multirow{-2}{*}{\textbf{Systems} (X $\rightarrow$ 69)}                        &   \multirow{-2}{*}{ara}              &    \multirow{-2}{*}{cmn}              &    \multirow{-2}{*}{eng}                       &  \multirow{-2}{*}{ind}               &       \multirow{-2}{*}{jpn}           &   \multirow{-2}{*}{kor}                                        &         \multirow{-2}{*}{tha}         &         \multirow{-2}{*}{tur}        &       \multirow{-2}{*}{vie}         &      \multirow{-2}{*}{\textbf{Avg.}}               \\
\midrule
  \multicolumn{11}{c}{\textbf{Cascaded ASR+MT Models}} \\
\midrule
Whisper-Large-V3 + NLLB-200-3.3B~\cite{nllb2024scaling}      &  21.0&18.0&30.1&23.4&18.5&19.3&16.8&23.1&18.4&21.0     \\   
Whisper-Large-V3 + LLaMAX3-8B-Alpaca~\cite{lu2024llamax}     &   15.7&14.4&23.5&18.9&14.2&15.1&12.6&17.2&14.6&16.2   \\          
\midrule
\multicolumn{11}{c}{\textbf{End-to-end S2TT Models}} \\
\midrule
SeamlessM4T-V2-Large~\cite{seamless2025joint}    & 15.8&13.5&31.7&15.4&11.9&14.3&11.7&16.0&14.3&16.1      \\
MCAT-Large-27B (ours)   & 19.0&16.8&30.3&23.0&17.1&19.7&14.8&22.8&17.1&20.1     \\
\bottomrule
\end{tabular}}
\label{0970_spbleu}
\end{table*}

\end{document}